%% file: bare_conf.tex
\documentclass[conference, a4paper]{IEEEtran}
\usepackage{comment}
\usepackage{color}
\usepackage{algorithm}
\usepackage{algpseudocode}
\usepackage{amsmath, amssymb}
\usepackage{multirow}
\usepackage{bbm}
\usepackage[switch,columnwise]{lineno}

\usepackage{booktabs}
\DeclareMathOperator*{\argmin}{arg\,min}
\DeclareMathOperator*{\argmax}{arg\,max}

\ifCLASSINFOpdf
   \usepackage[pdftex]{graphicx,xcolor}
   \usepackage[colorlinks,citecolor=green,urlcolor=blue,bookmarks=false,hypertexnames=true]{hyperref} 
   \usepackage{subfigure}

\else
\fi
\begin{document}
%

\title{Subspace-Guided Feature Reconstruction for Unsupervised Anomaly Localization}

\author{\IEEEauthorblockN{Katsuya Hotta$^1$,
Chao Zhang$^2$, Yoshihiro Hagihara$^1$, and Takuya Akashi$^3$}

\IEEEauthorblockA{$^1$Iwate University, 4-3-5 Ueda, Morioka-shi, Iwate, 020-8551, Japan\\
$^2$University of Toyama, 3190 Gofuku, Toyama-shi, Toyama, 930-8555, Japan\\
$^3$Okayama University, 3-1-1 Tsushima-naka, Kita-ku, Okayama, 700-8530, Japan}
Email: hotta@iwate-u.ac.jp}

\maketitle
\thispagestyle{plain}
\pagestyle{plain}
\begin{abstract}
Unsupervised anomaly localization aims to identify anomalous regions that deviate from normal sample patterns. Most recent methods perform feature matching or reconstruction for the target sample with pre-trained deep neural networks. However, they still struggle to address challenging anomalies because the deep embeddings stored in the memory bank can be less powerful and informative. Specifically, prior methods often overly rely on the finite resources stored in the memory bank, which leads to low robustness to unseen targets. In this paper, we propose a novel subspace-guided feature reconstruction framework to pursue adaptive feature approximation for anomaly localization. It first learns to construct low-dimensional subspaces from the given nominal samples, and then learns to reconstruct the given deep target embedding by linearly combining the subspace basis vectors using the self-expressive model. Our core is that, despite the limited resources in the memory bank, the out-of-bank features can be alternatively ``mimicked'' to adaptively model the target. Moreover, we propose a sampling method that leverages the sparsity of subspaces and allows the feature reconstruction to depend only on a small resource subset, contributing to less memory overhead. Extensive experiments on three benchmark datasets demonstrate that our approach generally achieves state-of-the-art anomaly localization performance.
\end{abstract}


%
\IEEEpeerreviewmaketitle

\input{paper/introduction}

\input{paper/relatedwork}

\input{paper/methodology}

\input{paper/experiment}

\input{paper/conclusion}


\bibliographystyle{IEEEtran}
\bibliography{myref}


\end{document}

%% file: paper/introduction.tex
\section{Introduction}
\label{sec:sec1}
Anomaly localization task for visual inspection, which refers to detecting visual anomalies in images that deviate from normal sample patterns, plays a key role in many practical applications such as product quality control~\cite{bergmann2022beyond,zou2022spot,shah2023two} and video surveillance~\cite{li2013anomaly,liu2018future,chen2023spatial}.
In practical scenarios, anomalies are rare in production lines, rendering the acquisition of a suitable dataset for training supervised methods impractical.
To tackle this problem, recent approaches have explored one-class classification (OCC) anomaly localization, leveraging only abundantly available anomaly-free images (see~\cite{tao2022deep} for a review).
Early works based on OCC generally train autoencoders (AEs) to reconstruct normal images only.
For an arbitrary testing image, the poorly reconstructed regions localize anomalies. As such, the models with higher generalization capacity would by contrast degrade the performance.
This makes later efforts focus on tuning the generalization strength of models, such as exploring the bottleneck layer for reconstruction.

\begin{figure}[t]
\centering
\includegraphics[width=1.\linewidth]{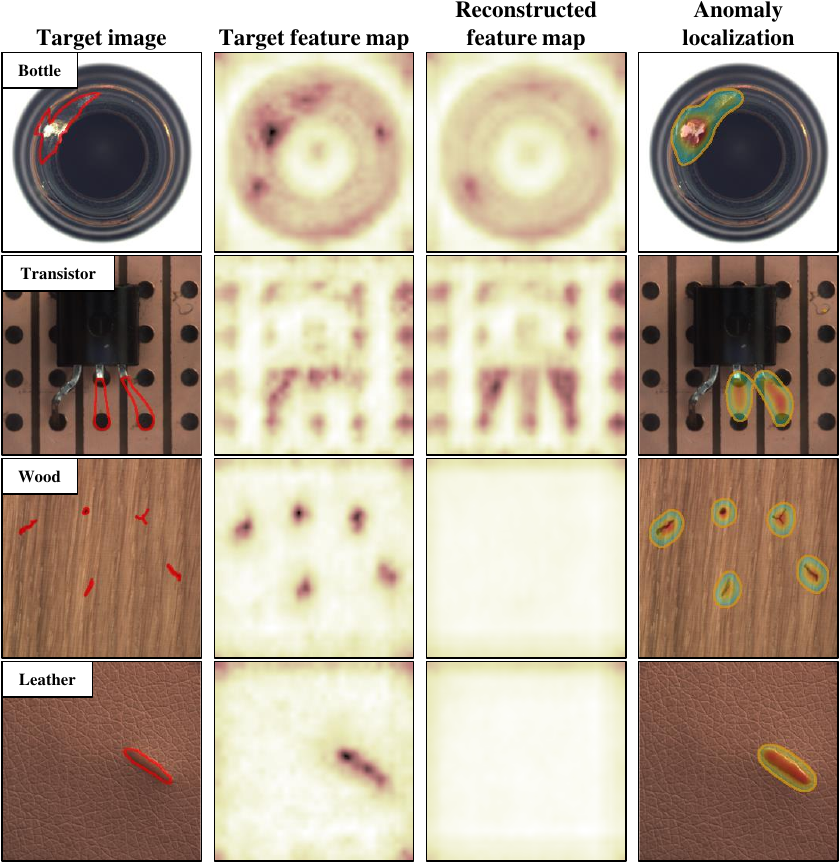}
\caption{An example of anomaly localization results from our method on the MVTec AD benchmark dataset. The \textcolor{red}{red} boundary in the target image indicates anomalous regions. Each feature map is shown in \textcolor{red}{red} color gradients for visualization. Anomaly localization results are shown by \textcolor{orange}{orange} for anomaly boundaries and \textcolor{blue}{blue}-\textcolor{red}{red} color gradients for anomaly intensity.}
  \label{fig:fig1}
\end{figure}

Modern OCC anomaly localization methods~\cite{deng2022anomaly,liu2023simplenet} leverage the knowledge of pre-trained deep neural networks for ImageNet~\cite{deng2009imagenet} classification.
In a pioneering study, \cite{cohen2020sub} attempt to extract features exclusively from normal data (i.e., nominal data) by utilizing a pre-trained neural network and subsequently keep these features in a memory bank.
Identifying abnormal regions that deviate from the nominal patterns stored in the memory bank is facilitated through feature matching techniques.
This concept of using feature matching lays the groundwork for state-of-the-art methods~\cite{roth2022towards, sun2023mbmf} to achieve strong spatial localization of anomalies despite the lack of adaptation to specific feature distributions.
However, as pointed out in \cite{roth2022towards, lee2022cfa}, feature matching using deep feature embeddings is a non-adaptive approach.
Non-adaptive approaches have limited confidence in matching unseen features because they rely heavily on the finite resources stored in a memory bank.
Although this limitation can be alleviated by preparing a vast number of feature patterns, as memory bank resources increase, the anomaly localization process becomes computationally expensive, and latency increases.
One straightforward strategy to address this challenge is to screen the data before storing them in a memory bank. 
Given the constraints imposed by computational and memory limitations, \cite{cohen2020sub} and \cite{roth2022towards} employ pre-selections based on kNN~\cite{eskin2002geometric} and coreset subsampling~\cite{agarwal2005geometric}, respectively. 
In contrast, \cite{liu2023simplenet} attempts to generate synthetic anomalies in a pre-trained feature space to improve the discriminative performance for localizing anomalous features.
While these approaches enhance the feature information extracted from nominal data during inference, they do not inherently tackle adapting to the feature distribution of specific data.
Thus, approaches that represent specific feature distributions have shifted to a policy of reconstructing target features.
Although conventional methods~\cite{bergmann2018improving,bergmann2019mvtec, yang2020dfr} can represent a wider range of normal appearances, they suffer from a shortcut learning problem in which even abnormal regions are reconstructed accurately due to their high generalization ability.

In this paper, we propose a novel subspace-guided framework to provide a better solution to anomaly localization. 
Our method aims to pursue a more reliable feature approximation for unseen targets by learning low-dimensional subspaces~\cite{hotta2022component, hotta2023pmssc}, which can be easily derived using the nominal sample features in the memory bank. 
Specifically, by introducing the approximation policy of the self-expressive model~\cite{elhamifar2013sparse}, the unseen target features can be compactly represented by a linear combination of features from nominal samples lying on one specific underlying subspace. 
As such, despite the lack of target information in the memory bank, the target feature can be readily produced by resorting to the ideal subspace.
Moreover, our framework addresses anomaly localization by leveraging the insight that as our subspace-guided approximation strategy only enables reproducing the anomalous-free features, the anomalous regions would induce a significant reconstruction error.
We thus define the anomaly score as the difference between the target feature and the reconstructed feature via our method.
Compared to prior approaches using feature matching, which relies largely on finite resources stored in a memory bank, our subspace-guided feature reconstruction yields a richer data coverage and robustness to challenging anomalies.
Ideally, fully utilizing the storage in the memory bank would lead to a strong target feature approximation capacity. 
Nevertheless, it would cause a prohibitively large computational overhead. 
To address this, we leverage the sparsity~\cite{donoho2006most} of subspaces and further propose a sampling method that allows the feature reconstruction to depend only on a small resource subset, improving computational efficiency.
As shown in Fig.~\ref{fig:fig1}, our method reconstructs features to mimic a given target sample to localize anomalies during inference.
Extensive experiments on three public datasets demonstrate that our approach generally achieves state-of-the-art anomaly localization performance.

In summary, our paper includes the following main contributions: 
\begin{itemize}
    \item We propose a novel anomaly localization framework that incorporates the subspace-guided feature reconstruction mechanism. It adaptively mimics the out-of-memory bank features by linearly combining the features in one specific subspace derived from the given nominal samples to encourage a sufficient data coverage.
    \item We introduce a sparsity-inspired sampling technique for subspace basis vectors to mitigate memory expense in reconstructing target features.
    \item We perform extensive experiments to demonstrate that our method yields state-of-the-art anomaly localization accuracy with only very few nominal data.
\end{itemize}

\begin{figure*}[t]
\centering
    \includegraphics[width=0.95\linewidth]{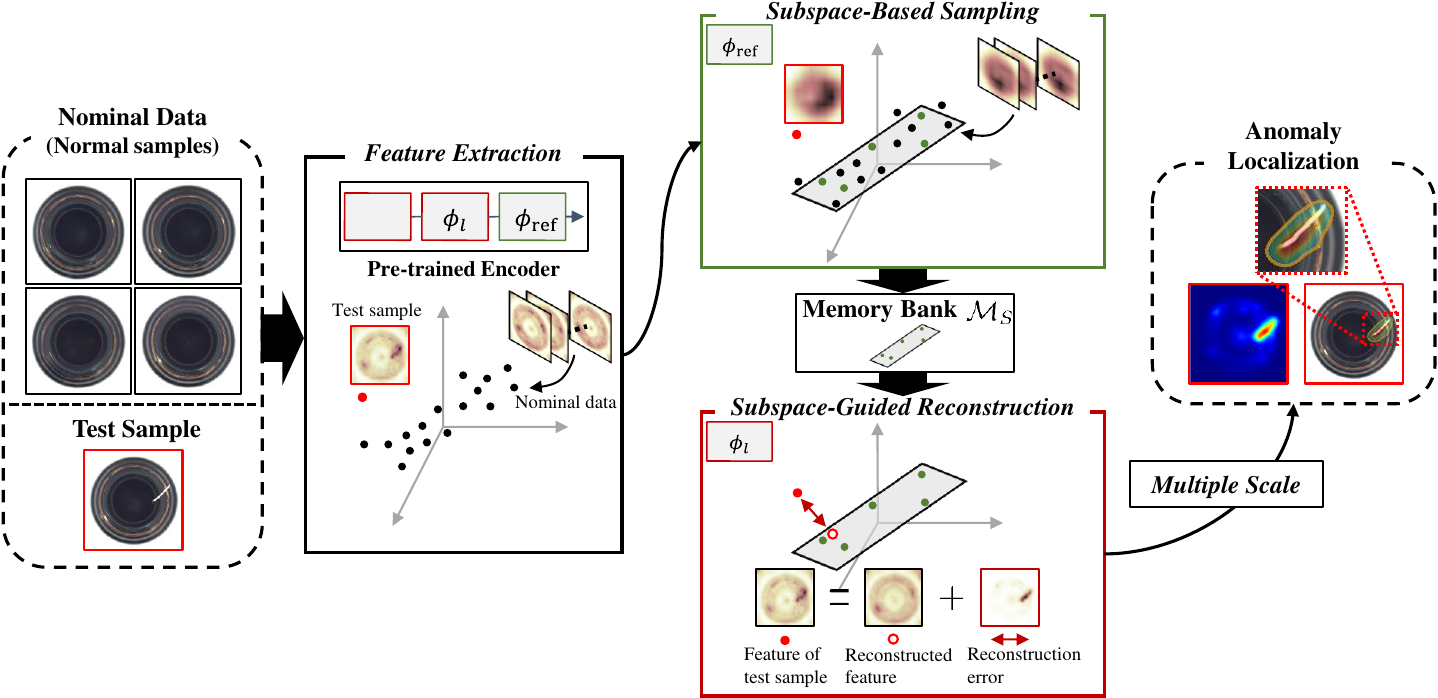}
  \caption{Overview of our approach. Features are extracted from nominal data $\mathcal{D}$ and a test sample $J$ through a pre-trained network $\phi$. To achieve reduced memory expense and computational complexity, features in hierarchy level $l_{\mathrm{ref}}$ are utilized to estimate the subspace in which the nominal data lie. Only a limited number of data, sufficient for recovering this subspace, is sampled and stored in the memory bank $\mathcal{M}_{S}$. By exclusively utilizing the data in $\mathcal{M}_{S}$, the reconstruction methodology based on the self-expressive model is employed to identify anomalous regions in the test sample. The proposed method performs pixel-level anomaly localization by scoring at multiple hierarchies to benefit from deeper features.
  }
  \label{fig:fig2}
\end{figure*}

%% file: paper/relatedwork.tex
\section{Related Work}
\label{sec:sec2}
Anomaly localization focuses on identifying visually anomalous regions in an image. This task is more challenging than anomaly detection~\cite{de2022hybrid,bergman2020deep,rudolph2022fully}, which distinguishes patterns in data that deviate from expected behavior.
In the context of recent unsupervised anomaly localization methods, one of the key issues is how to efficiently identify anomalous regions during inference based on the nominal dataset available beforehand.

Early literature~\cite{akcay2019ganomaly, akccay2019skip} involves reconstructing normal images from anomalous images using an encoder-decoder network.
The anomaly localization capability of these methods is based on the assumption that the trained network cannot accurately reconstruct anomalous regions, and anomalies can be identified by comparing the target image with its reconstructed image.
In particular, AE$_{SSIM}$~\cite{bergmann2018improving} introduces a structural similarity metric into the AE model~\cite{kingma2013auto} to capture the structural information of the image, thus enhancing the anomaly localization ability of the model.
However, these methods tend to accurately reconstruct anomalous regions due to their high generalization ability, leading to false negatives in anomaly localization as the complexity of the problem increases.
To address this issue, methods starting with~\cite{bergman2020deep} employ pre-trained networks on extensive external natural image datasets like ImageNet~\cite{deng2009imagenet}, capable of representing fine-grained visual features.

Recent methods, which introduce the innovative concept of utilizing nominal feature representations extracted by these pre-trained models, have significantly improved anomaly localization performance compared to earlier approaches.
\cite{bergmann2020uninformed} proposed a teacher-student framework based on knowledge distillation~\cite{hinton2015distilling}.
In this framework, during training, the student learns only from the knowledge of nominal samples transferred by the teacher. 
Following this knowledge distillation framework, STFPM~\cite{wang2021student} introduces a feature pyramid matching mechanism that alleviates the incompleteness of transferred knowledge loss by matching corresponding multi-scale features in the student-teacher network.
RD~\cite{deng2022anomaly} proposes an inverse distillation approach that extends the teacher-student framework to enhance the diversity of anomaly representations.
Several studies propose training anomaly localization models in a self-supervised manner by simulating pseudo-anomalous data~\cite{capogrosso2024diffusion}.
CutPaste~\cite{li2021cutpaste} generates synthetic anomalies using a simple technique that involves cutting patches from an image and placing them at random locations within the image.
DeSTSeg~\cite{zhang2023destseg} uses two-dimensional Perlin noise~\cite{zavrtanik2021draem, zavrtanik2022dsr} to generate more realistic anomalous images.
Yao et al.~\cite{yao2024few} proposed an anomaly localization framework based on edge enhancement and cascade optimization of residual features to generate augmented images that closely resemble the distribution of real anomalous images.
DiffAD~\cite{zhang2023unsupervised} leverages the diffusion model~\cite{rombach2022high} to capture structural changes in images, achieving strong performance in anomaly localization.
Similarly, RealNet~\cite{zhang2024realnet} introduces a strength-controllable diffusion-based anomaly synthesis strategy to generate realistic yet diverse anomaly samples that closely align with natural data distributions.

Reconstruction-based approaches using deep embeddings have also been proposed recently.
Unlike AE$_{SSIM}$, DFR~\cite{yang2020dfr} employs the deep features extracted by a pre-trained network as the reconstruction target. 
DSR~\cite{zavrtanik2022dsr} utilizes the latent space of pre-trained autoencoders to generate synthetic anomalies, enabling the generation of anomaly regions at the feature level.
It is worth noting that the aforementioned approaches demand a large number of training data and necessitate periodic retraining whenever the training dataset is updated. 
In contrast, anomaly localization approaches based on the memory bank framework, which are the most relevant to our proposed method, do not require a training step. 
This characteristic is attractive for practical applications in industrial anomaly localization.

A representative work of the memory bank-based approach is SPADE~\cite{cohen2020sub}, which achieves anomaly localization by storing nominal features extracted from a pre-trained neural network in a memory bank and then performing feature matching.
While storing all nominal data in the memory bank can improve anomaly localization performance, it may not be practical due to computational constraints.
To enhance the computational efficiency of this feature matching process, SPADE incorporates pre-selection using kNN~\cite{eskin2002geometric}.
However, maintaining accuracy while using kNN-based pre-selection remains computationally demanding, mainly due to the limited number of reductions permissible within the memory bank.
Consequently, subsequent methods have focused on maximizing the information of nominal features retained in memory banks.
PaDiM~\cite{defard2021padim} adopts a strategy by evaluating the Mahalanobis distance~\cite{mahalanobis2018generalized} at the patch level instead of using the time-intensive kNN method.
PatchCore~\cite{roth2022towards} proposes to build a memory bank that considers the negative effects of biased features of pre-trained networks on anomaly localization.
A crucial limitation of these methods lies in their reliance on feature matching in the memory bank for anomaly localization without fundamentally addressing the task of adapting to the feature distribution of specific data.

Following this line of research, our approach introduces the concept of subspaces~\cite{vidal2011subspace} into the task of anomaly localization, thereby affording two advantages: i) an adaptive feature representation guided by subspace-preserving properties; ii) memory bank reduction motivated by sparsity.
Compared to prior approaches, our framework is less reliant on features stored in a memory bank and instead explores faithful approximation of normal features through subspace structures that compactly compress patterns of nominal features. 
This allows the model to reconstruct normal features while excluding abnormal regions in unseen targets and significantly alleviates the issue of shortcut learning, which is widely confirmed in conventional reconstruction-based models.
Moreover, the memory bank reduction motivated by sparsity can reduce the reconstruction burden while maintaining performance by selectively retaining a small part of data that accurately represents nominal feature patterns.

%% file: paper/methodology.tex
\section{Anomaly Localization via Subspace Learning}
\label{sec:sec3}
In this section, we propose a framework that utilizes a self-expressive model for subspace learning to arrive at localization decisions, as illustrated in Fig.~\ref{fig:fig2}.
To this end, we present an algorithm designed to capture the similarity between a target feature and an approximated feature by a self-expressive model (Sec.~\ref{sec:sec3_1}).
We then explicate the sampling technique that employs a subspace structure to decrease the inference time (Sec.~\ref{sec:sec3_2}), and we finally summarize the comprehensive algorithm up to the localization of anomalies (Sec.~\ref{sec:sec3_3}).

\subsection{Subspace-Guided Feature Reconstruction}
\label{sec:sec3_1}
We first describe our policy for localizing anomalous regions by approximating the features using a self-expressive model.
Formally, we define the nominal data containing only normal images as $\mathcal{D} = \{ I_{1}, \cdots, I_{n}, \cdots, I_{N}\}$, and the test image provided during inference, which can be either normal or abnormal, is denoted by $J$.
Each image is characterized by its width $w_{\ast}$, height $h_{\ast}$, and number of channels $c_{\ast}$ (i.e., $I_{n} \in \mathbb{R}^{w_{\ast}\times h_{\ast} \times c_{\ast}}$ and $J \in \mathbb{R}^{w_{\ast}\times h_{\ast}\times c_{\ast}}$).
Our approach leverages a powerful feature extractor, capable of capturing high-quality deep features from images, facilitating feature reconstruction at the pixel level.
To realize this, as suggested in~\cite{bergman2020deep, cohen2020sub, roth2022towards}, we employ pre-trained deep neural networks on ImageNet~\cite{deng2009imagenet} as feature extractors. 
This network captures meaningful features at different hierarchies, with each hierarchical feature playing an important role.
As such, at the $l$-th hierarchical level of the pre-trained network $\phi$, we denote the feature map $\phi_{l}(I_{n})$ for image $I_{n} \in \mathcal{D}$, and the feature map $\phi_{l}(J)$ for image $J$.
The hierarchical level $l$ refers to the feature maps indexed from ResNet-like~\cite{he2016deep} architectures, where $l \in \{1,2,3,4\}$ denotes the final output of each spatial resolution block (i.e., $\phi_{l}(I_{n}) \in \mathbb{R}^{w_{l} \times h_{l} \times c_{l}}$ and $\phi_{l}(J) \in \mathbb{R}^{w_{l} \times h_{l} \times c_{l}}$).

We propose a feature reconstruction mechanism by introducing a self-expressive model that exhibits enhanced feature representation capabilities compared to feature matching techniques that rely on limited resources of $\mathcal{D}$.
We extend the previously introduced notation to formulate a feature reconstruction at the pixel level.
The feature map $\phi_{l}(I_{n})$ is defined as the feature vector $\boldsymbol{x}_{n,l} \in \mathbb{R}^{w_{l}h_{l}c_{l}}$ obtained by concatenating its dimensions into a $w_{l}h_{l}c_{l}$-dimensional representation. 
Similarly, $\phi_{l}(J)$ is defined as $\boldsymbol{y}_{l} \in \mathbb{R}^{w_{l}h_{l}c_{l}}$.
Given a data matrix $X_{l}=[\boldsymbol{x}_{1,l},\boldsymbol{x}_{2,l},\cdots,\boldsymbol{x}_{N,l}] \in \mathbb{R}^{w_{l}h_{l}c_{l} \times N}$ comprising column-wise nominal feature vectors $\boldsymbol{x}_{n,l}$, the reconstruction of the test feature vector in the self-expressive model can be described as 
\begin{equation}
    \boldsymbol{y}_{l} = X_{l}\boldsymbol{c}_{l} + \boldsymbol{e}_{l},
    \label{eq:eq1}
\end{equation}
where $\boldsymbol{c}_{l} \in \mathbb{R}^{N}$ is a coefficient vector and $\boldsymbol{e}_{l} \in \mathbb{R}^{w_{l}h_{l}c_{l}}$ is an anomalous term.
$\boldsymbol{e}_{l}$ refers to an out-of-distribution feature that exhibits the difference from the distribution of nominal features in $X_{l}$.
In an ideal context, if $\boldsymbol{y}_{l}$ follows the distribution of $X_{l}$ (i.e., when $\boldsymbol{y}_{l}$ represents normal data), the anomaly term should be precisely zero. Conversely, if $\boldsymbol{y}_{l}$ deviates from this distribution (i.e., when $\boldsymbol{y}_{l}$ contains anomalous regions), it should be non-zero.
In the optimization of Eq.~(\ref{eq:eq1}), feasible solutions are generally non-unique because the number of data that lie in a subspace is typically larger than its dimensionality.
Nonetheless, at least one $\boldsymbol{c}_{l}$ exists where $c_{i,l}$ is non-zero only if the data $\boldsymbol{x}_{i,l}$ and $\boldsymbol{y}_{l}$ lie in the same subspace, which is referred to as subspace-preserving~\cite{you2015geometric, you2016scalable}.
It is worth mentioning that nominal features that satisfy the subspace-preserving property can represent only normal features modeled in the same low-dimensional subspace. Thus, unseen features would cause large reconstruction errors to indicate anomalies, which constitutes the core advantage of introducing low-dimensional subspaces in our framework. 
\textit{In summary, our method is conceptually different from prior approaches where the models simply reconstruct a copy of target features without considering the data structure of the nominal distribution.}

\begin{algorithm}[t]
\caption{Optimization for Solving Problem (\ref{eq:eq2})}         
\label{alg:alg1}  
\begin{algorithmic}[1]
\renewcommand{\algorithmicrequire}{\textbf{Input:}}
\renewcommand{\algorithmicensure}{\textbf{Output:}}
  \Require{$\mathcal{M}$, $X_{l}$, $\boldsymbol{y}_{l}$, $s$, and $\epsilon$}
  \State{Initialize $k=0$, anomalous term $\boldsymbol{e}_{l} = \boldsymbol{y}_{l}$, and $\mathcal{S}=\emptyset$;}
  \While{$k < s$ and $\|\boldsymbol{e}_{l}\|_{2} > \epsilon$}
  \State{Find $j^{\ast}$ via Eq. (\ref{eq:eq3});}
  \State{Update $\mathcal{S} \leftarrow \mathcal{S} \cup \{j^{\ast}\}$;}
  \State{Estimate $\boldsymbol{c}_{l}$ via Eq. (\ref{eq:eq4});}
  \State{Update $\boldsymbol{e}_{l}$  via Eq. (\ref{eq:eq5});}
  \State{$k = k + 1$;}
  \EndWhile{}
  \Ensure{Coefficient vector $\boldsymbol{c}_{l}$ and anomalous term $\boldsymbol{e}_{l}$}
\end{algorithmic}
\end{algorithm}

\noindent\textbf{Optimization.\hspace{2px}}
The coefficient vector $\boldsymbol{c}_{l}$ in Eq.~(\ref{eq:eq1}), which satisfies the subspace-preserving representations, can be solved by considering the problem of minimizing the anomalous term as follows:
\begin{equation}
    \min_{\boldsymbol{c}_{l}}\| \boldsymbol{y}_{l} - X_{l}\boldsymbol{c}_{l} \|_{2}^{2} \quad \mathrm{s.t.} \quad \| \boldsymbol{c}_{l}\|_{0} \leq s,
    \label{eq:eq2}
\end{equation}
where $|\cdot|_{0}$ represents the $\ell_{0}$ pseudo-norm counting the number of non-zero entries in the vector, and $s$ is a tuning parameter that controls the sparsity of the solution by selecting up to $s$ entries in the coefficient vector $\boldsymbol{c}_{l}$.
We utilize the orthogonal matching pursuit (OMP) algorithm~\cite{pati1993orthogonal, tropp2004greed, davenport2010analysis} to solve this optimization problem and recover a subspace-preserving solution.

To efficiently solve the Eq.~(\ref{eq:eq2}), we introduce Algorithm~\ref{alg:alg1} based on the OMP algorithm.
Formally, we commence by initializing the support set $\mathcal{S}$ and the anomalous term $\boldsymbol{e}_{l}$ as $\mathcal{S} = \emptyset$ and  $\boldsymbol{e}_{l} = \boldsymbol{y}_{l}$, respectively.
Subsequently, at each iteration, we update the index set $\mathcal{S}$ by adding one index $j^{\ast}$, which is computed as follows:
\begin{equation}
    j^{\ast} = \argmax_{j \in \mathcal{M} \setminus \mathcal{S}} \boldsymbol{x}_{j,l}^{\top} \boldsymbol{e}_{l},
    \label{eq:eq3}
\end{equation}
where the memory bank $\mathcal{M} \in \{1,\cdots,N\}$ is the candidate index set of the nominal features for reconstruction.
After updating $\mathcal{S}$ with $j^{\ast}$ (i.e., $\mathcal{S} \leftarrow \mathcal{S} \cup \{j^{\ast}\}$), we solve the following problem:
\begin{equation}
     \min_{\boldsymbol{c}_{l}} \| \boldsymbol{y}_{l} - X_{l} \boldsymbol{c}_{l}\|_{2}^{2} \hspace{10px} \mathrm{s.t.} \hspace{3px} \mathrm{supp}(\boldsymbol{c}_{l}) \subseteq \mathcal{S}.
    \label{eq:eq4}
\end{equation}
where $\mathrm{supp}(\cdot)$ is the support function that returns the subgroup of the domain containing elements not mapped to zero.
By utilizing the minimized coefficient vector $\boldsymbol{c}_{l}$ in iteration $k$, we can update the anomalous term $\boldsymbol{e}_{l}$ through the estimation of the approximation error between $\boldsymbol{y}_{l}$ and the reconstructed feature $X_{l}\boldsymbol{c}_{l}$:
\begin{equation}
    \boldsymbol{e}_{l} \leftarrow \boldsymbol{y}_{l} - X_{l} \boldsymbol{c}_{l}.
    \label{eq:eq5}
\end{equation}
We repeat the optimization process in Eq.~(\ref{eq:eq3})$\sim$(\ref{eq:eq5}) until either the number of iterations $k$ reaches the limit $s$ or $\boldsymbol{e}_{l}$ is smaller than the threshold $\epsilon$.

\begin{figure}[t]
    \centering
    \subfigure[]{\includegraphics[width=0.31\linewidth]{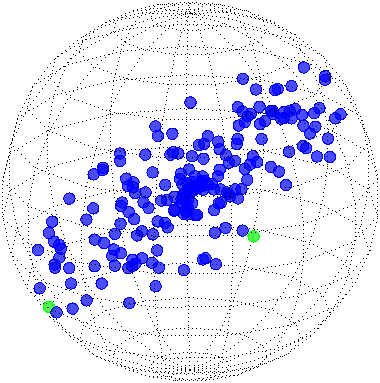}
    \label{fig:3a}
    }
    \subfigure[]{\includegraphics[width=0.31\linewidth]{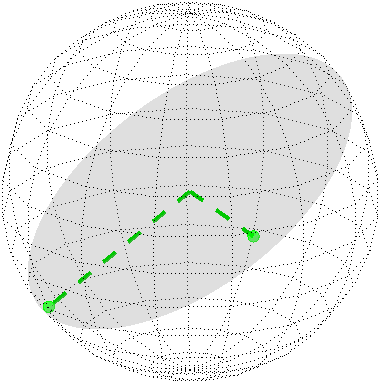}
    \label{fig:3b}
    }
    \subfigure[]{\includegraphics[width=0.31\linewidth]{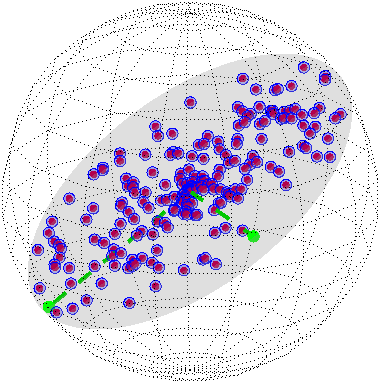}
    \label{fig:3c}
    }
    \caption{
    Illustrative example of covering out-of-bank data ($\mathcal{M} \setminus \mathcal{M}_{S}$) from a few data ($\mathcal{M}_{S}$) chosen through subspace-based sampling in a toy dataset: consider a synthetic dataset created by randomly generating data points lying on a single subspace of $\mathbb{R}^{2}$ in the ambient space of $\mathbb{R}^{3}$. The proposed subspace-based sampling strategy selects only very few basis vectors (\textcolor{green}{green}). Through a linear combination of these data points, a comprehensive approximation (\textcolor{red}{red}) is achieved for the out-of-bank data (\textcolor{blue}{blue}) that lie in the same underlying subspace (\textcolor{gray}{gray plane}).
    }
    \label{fig:3}
\end{figure}

\subsection{Subspace-Based Sampling Technique for Memory Bank Reduction}
\label{sec:sec3_2}
Although our model, as described in Sec.~\ref{sec:sec3_1}, is capable of predicting anomalous regions at the pixel level, utilizing a self-expressive model for reconstruction entails notable memory expense.
Moreover, it substantially increases inference time, particularly when dealing with an extensive volume of high-dimensional data.
We introduce a subspace-based sampling technique into our model to address this issue, motivated by~\cite{cohen2020sub,roth2022towards}.
This technique alleviates the reconstruction burden while maintaining performance by selectively retaining a limited number of in-sample data within the memory bank $\mathcal{M}$, which effectively approximates the out-of-bank data.
Specifically, our approach aims to identify $\mathcal{M}$-subspace $\mathcal{M}_{S}$ ($\mathcal{M}_{S} \in \mathcal{M}$) composed exclusively of entries that satisfy the subspace-preserving properties by selecting a few basis vectors from the subspace in the hierarchy $\mathrm{ref}=l_{\mathrm{ref}}$ ($l < l_{\mathrm{ref}}$), which represents high-level features:
\begin{equation}
\begin{split}
    \boldsymbol{c}_{\mathrm{ref}}^{\ast} = &
    \argmin_{\boldsymbol{c}_{\mathrm{ref}}} \| \boldsymbol{y}_{\mathrm{ref}} - X_{\mathrm{ref}}\boldsymbol{c}_{\mathrm{ref}} \|_{2}^{2} \\
  & \quad \mathrm{s.t.} \quad \| \boldsymbol{c}_{\mathrm{ref}}\|_{0} \leq s_{\mathrm{ref}},
\end{split}
\label{eq:eq6}
\end{equation}

\begin{equation}
    \mathcal{M}_{S} = \mathrm{supp}(\boldsymbol{c}_{\mathrm{ref}}^{\ast}),
    \label{eq:eq7}
\end{equation}
where $s_{\mathrm{ref}}$ is a tuning parameter that controls the sampling rate.
The solution of Eq.~(\ref{eq:eq6}) satisfying the subspace-preserving property can be solved by Algorithm~\ref{alg:alg1} as in Eq.~(\ref{eq:eq2}). 

The proposed sampling technique differs significantly from \cite{cohen2020sub, roth2022towards} in that its ability to effectively represent the out-of-bank features contained within $\mathcal{M}$ through the reconstruction process.
The indices encompassed within $\mathcal{M}_{S}$ serve as basis vectors that form a specific subspace.
In other words, the out-of-bank features $\boldsymbol{x}_{j,\mathrm{ref}}, \forall j \in \mathcal{M} \setminus \mathcal{M}_{S}$ lying in the subspace derived from $\mathcal{M}_{S}$ can be expressed as a linear combination of basis vectors that are in-sample data $\boldsymbol{x}_{i,\mathrm{ref}} \in \mathcal{M}_{S}$:
\begin{equation}
    \boldsymbol{x}_{j,\mathrm{ref}} = \sum_{i \in \mathcal{M}_{S}} \alpha_{i} \boldsymbol{x}_{i,\mathrm{ref}},
    \label{eq:eq8}
\end{equation}
where $\alpha_{i}$ represents an arbitrary coefficient.
Fig.~\ref{fig:3} provides a visual conceptual example that demonstrates a data coverage of out-of-bank using in-sample data in a toy dataset.
This expressiveness based on the subspace-preserving property facilitates a richer data coverage of the memory bank while concurrently mitigating the computational complexity associated with the reconstruction process (see experiments done in Sec.~\ref{sec:sec4-3-1}).

\begin{algorithm}[t]
\caption{Anomaly Localization via Learning Subspace Representations}         
\label{alg:alg2}  
\begin{algorithmic}[1]
\renewcommand{\algorithmicrequire}{\textbf{Input:}}
\renewcommand{\algorithmicensure}{\textbf{Output:}}
  \Require{Pre-trained network $\phi$, nominal dataset $\mathcal{D}$, test image $J$, parameters $l_{\mathrm{ref}}$, $l$, $s_{\mathrm{ref}}$, $s$, $\epsilon$;}
  \State{Set $\mathcal{M} \in \{1,\cdots,N\}$;}
  \Statex{$/^{\ast}$\quad \textit{Subspace-based sampling}~(Sec.~\ref{sec:sec3_2}) \quad$_{\ast}/$}
  \State{Given $\phi$, $l_{\mathrm{ref}}$, $\mathcal{D}$, and $J$, get $X_{\mathrm{ref}}$ and $\boldsymbol{y}_{\mathrm{ref}}$;}
  \State{Given $\mathcal{M}$, $X_{\mathrm{ref}}$, $\boldsymbol{y}_{\mathrm{ref}}$, and $s_{\mathrm{ref}}$, solve $\boldsymbol{c}_{\mathrm{ref}}^{\ast}$ via Algorithm~\ref{alg:alg1};}
  \State{Get $\mathcal{M}_{S}$ via Eq.~(\ref{eq:eq7})}
  \Statex{$/^{\ast}$\quad \textit{Subspace-guided feature reconstruction}~(Sec.~\ref{sec:sec3_1}) \quad$_{\ast}/$}
  \State{Given $\phi$, $l$, $\mathcal{D}$, and $J$, get $X_{l}$ and $\boldsymbol{y}_{l}$;}
  \State{Given $\mathcal{M}_{S}$, $X_{l}$, $\boldsymbol{y}_{l}$, and $s$, compute $\boldsymbol{e}_{l}$ via Algorithm~\ref{alg:alg1};}
  \State{Get anomaly scores from $\boldsymbol{e}_{l}$;}
  \Ensure{Anomaly score}
\end{algorithmic}
\end{algorithm}

\begin{figure}[t]
\centering
    \includegraphics[width=0.8\linewidth]{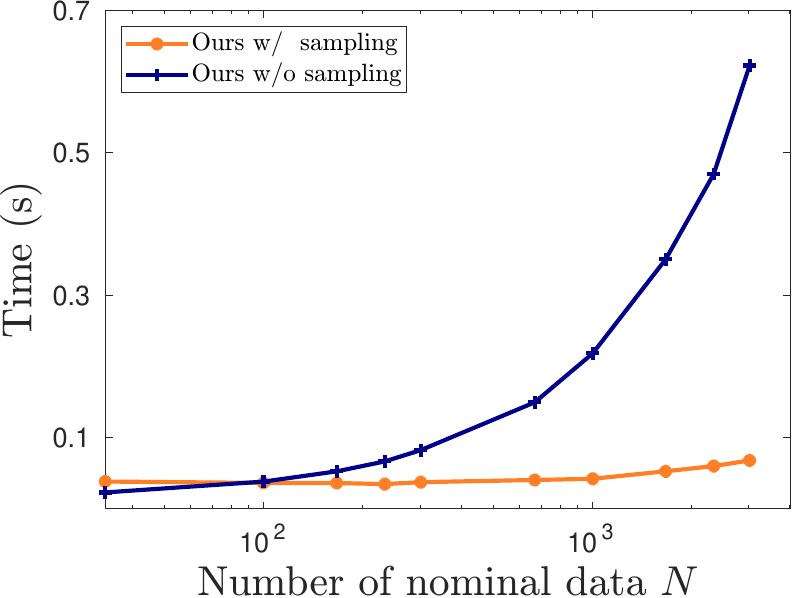}
  \caption{Curves of inference time with respect to the number of nominal data $N$, w/ (orange) or w/o (blue) our proposed sampling.}
  \label{fig:4}
\end{figure}

\subsection{Anomaly Localization via Learning Subspace Representations}
\label{sec:sec3_3}
We summarize the whole procedure of the proposed anomaly localization approach in Algorithm~\ref{alg:alg2}.
Given a test image $J$ and a nominal dataset $\mathcal{D}$ consisting of $N$ normal images, our objective is to identify anomalous features in $J$ that deviate from the feature patterns observed in $\mathcal{D}$.
Initially, we employ a network $\phi$ to extract $X_{\mathrm{ref}}$ and $\boldsymbol{y}_{\mathrm{ref}}$ from $\mathcal{D}$ and $J$, respectively.
We then employ the sampling technique described in Sec.~\ref{sec:sec3_2} to reduce the number of features stored in the memory bank $\mathcal{M}_{S}$ from $N$ to $s_{\mathrm{ref}}$.
Subsequently, similarly extracting the $l$-th hierarchical level features, we perform the subspace-guided feature reconstruction process, as described in Sec.~\ref{sec:sec3_1}, utilizing $\mathcal{M}_{S}$.
To maintain compatibility with the original input resolution, we resize $\boldsymbol{e}_{l}$ to match the size of feature map $\phi_{l}({J})$ and employ bi-linear interpolation for upscaling.
Furthermore, we apply a smoothing operation to the result using a Gaussian kernel with a width parameter $\sigma=4$. 
Note that our implementation does not optimize the specific value of $\sigma$.
Our algorithm is capable of utilizing anomaly scores across multiple scales (i.e., $l \in \{2,3,4\}$), similar to~\cite{cohen2020sub, wang2021student}.
For a comprehensive analysis, we kindly refer you to Sec.~\ref{sec:sec4-3}, which provides a detailed ablation study.

\begin{table*}[t]
\centering
\caption{Anomaly localization performance on MTD dataset~\cite{huang2020surface}. Comparative results are measured by AUROC and PRO.}
\label{tab:MTD}
\vspace{5pt}
\scalebox{1.}{ 
\begin{tabular}{lcccccccc}
\toprule
\textbf{Method}    & SPADE~\cite{cohen2020sub}  & STFPM~\cite{wang2021student}  & PaDiM~\cite{defard2021padim}  & PatchCore~\cite{roth2022towards} & DSR~\cite{zavrtanik2022dsr} & SimpleNet~\cite{liu2023simplenet} & DeSTSeg~\cite{zhang2023destseg} & Ours      \\ \hline
\textbf{AUROC}$\uparrow$     & 72.8   & 68.5   & 74.1   & \underline{84.8}      &57.7 & 78.9 & 74.0 & \textbf{85.7}      \\
\textbf{PRO}$\uparrow$       & 69.4   & 61.4   & 68.5   & \underline{72.3}      & 40.8 & 71.0 & \underline{72.3} & \textbf{73.1}      \\
\bottomrule
\end{tabular}
}
\end{table*}

\begin{table}[t]
\centering
\caption{Anomaly localization performance on BTAD dataset~\cite{mishra2021vt}. Comparative results are measured by AUROC / PRO.}
\label{tab:BTAD}
\vspace{5pt}
\scalebox{0.9}{ 
\begin{tabular}{lcccc}
\toprule
\textbf{Method}    & Case 01   & Case 02   & Case 03   & \textbf{Avg.}       \\ \hline
STFPM~\cite{wang2021student}     & 96.2 / 71.7 & \textbf{97.0} / 45.1 & \underline{99.3} / 96.4 & \textbf{97.5} / 71.0 \\
FastFlow~\cite{yu2021fastflow}    & \textbf{97.1} / 71.7 & 93.6 / \textbf{63.1} & 98.3 / 79.5 & 96.3 / 71.4 \\
PatchCore~\cite{roth2022towards} & \underline{97.0} / 64.9 & \underline{95.8} / 47.3 & 99.2 / 67.7 & \underline{97.4} / 60.0 \\
DSR~\cite{zavrtanik2022dsr} & 89.1 / 61.5 & 78.8 / 38.5 & 94.1 / 84.0 & 87.3 / 61.3 \\
SimpleNet~\cite{liu2023simplenet} & 96.2 / 69.6 & 95.0 / 53.3 & \textbf{99.5} / \underline{97.0} & 96.9 / 73.3 \\
DeSTSeg~\cite{zhang2023destseg} & 96.2 / \textbf{83.2} & 94.8 / 42.8 & \underline{99.3} / \textbf{97.2} & 96.8 / \underline{74.4} \\
Ours      & 96.2 / \underline{72.6} & 94.8 / \underline{54.7} & \underline{99.3} / \underline{97.0} & 96.8 / \textbf{74.8} \\
\bottomrule
\end{tabular}
}
\end{table}

\noindent\textbf{Complexity Analysis.\hspace{2px}}
Algorithm~\ref{alg:alg1} necessitates a certain number of inner products, which is contingent upon the size of the memory bank $\mathcal{M}$. 
Specifically, given a memory bank $\mathcal{M} \in \{1,\cdots,N \}$ containing all nominal data, $N$ inner products are required within a single iteration, resulting in the computational complexity of $\mathcal{O}(w_{l}h_{l}c_{l}Ns)$ for the optimization problem stated in Eq.~(\ref{eq:eq2}).
The sampling technique described in Sec.~\ref{sec:sec3_2} effectively reduces the number of indices in the memory bank from the original $N$ to $s_{\mathrm{ref}}$ with the computational complexity $\mathcal{O}(w_{\mathrm{ref}}h_{\mathrm{ref}}c_{\mathrm{ref}}Ns_{\mathrm{ref}})$.
Thus, the computational complexity of the subspace-guided feature reconstruction with $\mathcal{M}_{\mathrm{S}}$ obtained by applying the sampling technique is reduced to $\mathcal{O}(w_{l}h_{l}c_{l}s_{\mathrm{ref}}s)$.
Consequently, the overall computational complexity of our approach is $\mathcal{O}(w_{\mathrm{ref}}h_{\mathrm{ref}}c_{\mathrm{ref}}Ns_{\mathrm{ref}} + w_{l}h_{l}c_{l}s_{\mathrm{ref}}s)$.
In the case of WideResnet-50, the feature maps obtained from consecutive hierarchical levels follow the relationships $w_{l+1} = \frac{1}{2}w_{l}$, $h_{l+1} = \frac{1}{2}h_{l}$, $c_{l+1} = 2c_{l}$. 
By adhering to the memory bank reduction technique, we set $l_{\mathrm{ref}} = l+1$ and $s_{\mathrm{ref}} \simeq s \ll N$.
Based on these considerations, the computational complexity of our algorithm can be estimated as $\mathcal{O}(w_{\mathrm{ref}}h_{\mathrm{ref}}c_{\mathrm{ref}}Ns_{\mathrm{ref}} + w_{l}h_{l}c_{l}s_{\mathrm{ref}}s) = \mathcal{O}(w_{l}h_{l}c_{l}(\frac{1}{2}N+s)s_{\mathrm{ref}}) < \mathcal{O}(w_{l}h_{l}c_{l}Ns)$. 

Therefore, it is trivial to conclude that our sampling technique enjoys the advantage of an $N$-independent reconstruction process when using multi-scale anomaly scores. 
To experimentally study this property, we investigate the change of inference time with respect to the number of nominal data $N$, and plot the curve in Fig.~\ref{fig:4} by fixing $(w_{l},h_{l},c_{l},s_{\mathrm{ref}},s)$. 
It can be confirmed from the orange curve in Fig.~\ref{fig:4} that our sampling method contributes to an $N$-independent reconstruction, which generally aligns with our theoretical analysis. 
Further discussion regarding other parameters is presented in Sec.~\ref{sec:sec4-3-1}.

%% file: paper/experiment.tex
\begin{table*}[t]
\centering
\caption{Anomaly localization performance in terms of AUROC on MVTec AD dataset~\cite{bergmann2019mvtec}.}
\label{tab:MVT_AUROC}
\vspace{5pt}
\scalebox{0.8}{ 
\begin{tabular}{lcccccccccccccccc}
\toprule
\textbf{Method}       & Bottle        & Cable         & Capsule       & Carpet        & Grid          & Hazelnut      & Leather       & Metal Nut     & Pill          & Screw         & Tile          & Toothbrush    & Transistor    & Wood          & Zipper        & \textbf{Avg.} \\ \hline
AE$_{SSIM}$~\cite{bergmann2019mvtec} & 93.0          & 82.0          & 94.0          & 87.0          & 94.0          & 97.0          & 78.0          & 89.0          & 91.0          & 96.0          & 59.0          & 92.0          & 90.0          & 73.0          & 88.0          & 87.0          \\
SPADE~\cite{cohen2020sub}        & 98.4          & 97.2          & \textbf{99.0}    & 97.5          & 93.7          & \underline{99.1} & 97.6          & 98.1 & 96.5 & 98.9    & 87.4          & 97.9    & 94.1    & 88.5          & 96.5          & 96.0          \\
PaDiM~\cite{defard2021padim}        & 98.3          & 96.7    & 98.5          & \underline{99.1}          & 97.3    & 98.2          & 99.2          & 97.2          & 95.7          & 98.5          & 94.1          & 98.8          & \textbf{98.5}          & 94.9    & 98.5    & 97.5          \\
STFPM~\cite{wang2021student} & \underline{98.8}         & 95.5          & 98.3          & 98.8          & \underline{99.0}          & 98.5          & 99.3          & 97.6          & 97.8          & 98.3          & 97.4          & 98.9          & 82.5          & \textbf{97.2}          & 98.5          & 92.1          \\
PatchCore~\cite{roth2022towards}    & 98.6 & \textbf{98.5} & \underline{98.9} & \underline{99.1}    & 98.7 & 98.7          & 99.3 & 98.4          & 97.6    & \textbf{99.4} & 95.9    & 98.7          & \underline{96.4}          & 95.1          & \underline{98.9} & \textbf{98.1} \\
RIAD~\cite{zavrtanik2021reconstruction}    & 98.4 & 84.2 & 92.8 & 96.3    & 98.8 & 96.1          & 99.4 & 92.5          & 95.7    & 98.8 & 89.1    & 98.9          & 87.7          & 85.8          & 97.8 & 94.2 \\
DFR~\cite{yang2020dfr}    & 97.0 & 92.0 & \textbf{99.0} & 97.0    & 98.0 & 99.0          & 98.0 & 93.0          & 97.0    & 99.0 & 87.0    & 99.0          & 80.0          & 93.0          & 96.0 & 95.0 \\
DSR~\cite{zavrtanik2022dsr}    & 98.7 & 97.2 & 91.0 & 95.8    & \textbf{99.5} & 98.8      & 99.4 & 93.0          & 94.0    & 97.9 & \textbf{98.7}    & \underline{99.3}          & 78.0          & 91.6          & 98.2 & 95.4 \\
SimpleNet~\cite{liu2023simplenet} & 98.2 & \underline{97.9} & \textbf{99.0} & 98.2 & 98.7 & 98.2 & 99.1 & \underline{98.7} & \underline{98.7} & \underline{99.2} & 96.8 & 98.4 & 97.5 & 91.8 & 98.6 & \underline{97.9} \\
DeSTSeg~\cite{zhang2023destseg} & \textbf{99.2} & 96.2 & 98.5 & 95.0 & 98.6 & \textbf{99.6} & \textbf{99.8} & \textbf{98.8} & \textbf{99.2} & 97.4 & \underline{97.6} & \textbf{99.4} & 93.5 & \underline{96.1} & \textbf{99.1} & \underline{97.9} \\
Ours         & 98.6    & 97.5          & 98.5          & \textbf{99.2} & 95.9          & 98.7    & \underline{99.5}    &97.8    & 93.1          & 98.4          & 94.0 & \underline{99.3} & \textbf{98.5} & 95.6 & 98.7          & 97.6    \\ 
\bottomrule
\end{tabular}
}
\end{table*}

\begin{table*}[t]
\centering
\caption{Anomaly localization performance in terms of PRO on MVTec AD dataset~\cite{bergmann2019mvtec}.}
\label{tab:MVT_PRO}
\vspace{5pt}
\scalebox{0.8}{ 
\begin{tabular}{lcccccccccccccccc}
\toprule
\textbf{Method}       & Bottle        & Cable         & Capsule       & Carpet        & Grid          & Hazelnut      & Leather       & Metal Nut     & Pill          & Screw         & Tile          & Toothbrush    & Transistor    & Wood          & Zipper        & \textbf{Avg.} \\ \hline
AE$_{SSIM}$~\cite{bergmann2019mvtec} & 83.4          & 47.8          & 86.0          & 64.7          & 84.9          & 91.6          & 56.1          & 60.3          & 83.0          & 88.7          & 17.5          & 78.4          & 72.5          & 60.5          & 66.5          & 69.4          \\
SPADE~\cite{cohen2020sub}        & 95.5          & \underline{90.9}          & 93.7    & 94.7          & 86.7          & 95.4 & 97.2          & 94.4 & 94.6 & \underline{96.0}    & 75.6          & 93.5    & 87.4    & 87.4          & 92.6          & 91.7          \\
PaDiM~\cite{defard2021padim}        & 94.8          & 88.8    & 93.5          & 96.2          & 94.6    & 92.6          & 97.8          & 85.6          & 92.7          & 94.4          & 86.0          & 93.1          & 84.5          & \underline{91.1}    & 95.9    & 92.1          \\
STFPM~\cite{wang2021student} & 95.1         & 87.7          & 92.2          & 95.8          & \underline{96.6}          & 94.3          & 98.0          & \underline{94.5}          & \textbf{96.5}          & 93.0          & 92.1          & 92.2          & 69.5          & \textbf{93.6}          & 95.2          & 92.1          \\
PatchCore~\cite{roth2022towards}    & \underline{96.1} & \textbf{92.6} & \underline{95.5} & \underline{96.6}    & 95.9 & 93.9          & \underline{98.9} & 91.3          & 94.1    & \textbf{97.9} & 87.4    & 91.4          & 83.5          & 89.6          & \textbf{97.1} & \textbf{93.5} \\
DFR~\cite{yang2020dfr}    & 93.0 & 81.0 & \textbf{97.0} & 93.0    & 93.0 & \underline{97.0}          & 97.0 & 90.0          & 96.0    & \underline{96.0} & 79.0    & 93.0          & 79.0          & 91.0          & 90.0 & 91.0 \\
DSR~\cite{zavrtanik2022dsr}    & 94.8 & 85.3 & 82.9 & 93.4    & \textbf{98.4} & 92.1      & 97.7 & 89.8          & 92.5    & 89.8 & \textbf{97.0}    & 93.7          & 77.2          & 86.4          & 94.1 & 91.0 \\
SimpleNet~\cite{liu2023simplenet} & 92.6 & 90.8 & 92.9 & 93.1 & 94.1 & 89.8 & 97.2 & 87.8 & 94.1 & 95.9 & 92.7 & 91.3 & \underline{90.4} & 77.9 & 95.1 & 91.7 \\
DeSTSeg~\cite{zhang2023destseg} & \textbf{97.1} & 84.6 & 94.0 & 90.7 & 94.3 & \textbf{97.2} & \textbf{99.3} & \textbf{95.5} & \underline{96.4} & 87.8 & \underline{93.0} & \textbf{95.8} & 89.3 & 91.0 & \underline{96.6} & \textbf{93.5}          \\
Ours         & \underline{96.1}    & 89.2          & 93.6          & \textbf{97.1} & 85.7          & 95.4    & 98.7    & 92.7    & 88.4          & 92.2          & 87.6 & \underline{93.9} & \textbf{95.2} & \textbf{93.6} & 95.7          & \underline{93.0}    \\
\bottomrule
\end{tabular}
}
\end{table*}

\begin{table}[t]
\centering
\caption{Mean anomaly localization performance using AP per dataset collection.}
\label{tab:AP}
\vspace{5pt}
\scalebox{0.9}{ 
\begin{tabular}{lcccc}
\toprule
\textbf{Method} & MTD~\cite{huang2020surface} & BTAD~\cite{mishra2021vt} & MVTec AD~\cite{bergmann2019mvtec} & \textbf{Avg.} \\ \hline
SPADE~\cite{cohen2020sub} & 23.5 & 46.3 & 52.0 & 40.6 \\
STFPM~\cite{wang2021student} & 31.0 & \underline{51.9} & 49.8 & 44.2 \\
PatchCore~\cite{roth2022towards} & \underline{33.6} & 50.8 & 57.1 & 47.2 \\
DSR~\cite{zavrtanik2022dsr} & 8.9 & 21.2 & \underline{71.2} & 33.8  \\
SimpleNet~\cite{liu2023simplenet} & 33.0 & 41.9 & 50.1 & 41.7 \\
DeSTSeg~\cite{zhang2023destseg} & 31.5 & 39.1 & \textbf{73.2} & \underline{47.9} \\
Ours      & \textbf{34.1} & \textbf{53.0} & 63.1 & \textbf{50.1} \\
\bottomrule
\end{tabular}
}
\end{table}

\begin{figure*}[t]
\centering
    \includegraphics[width=0.90\linewidth]{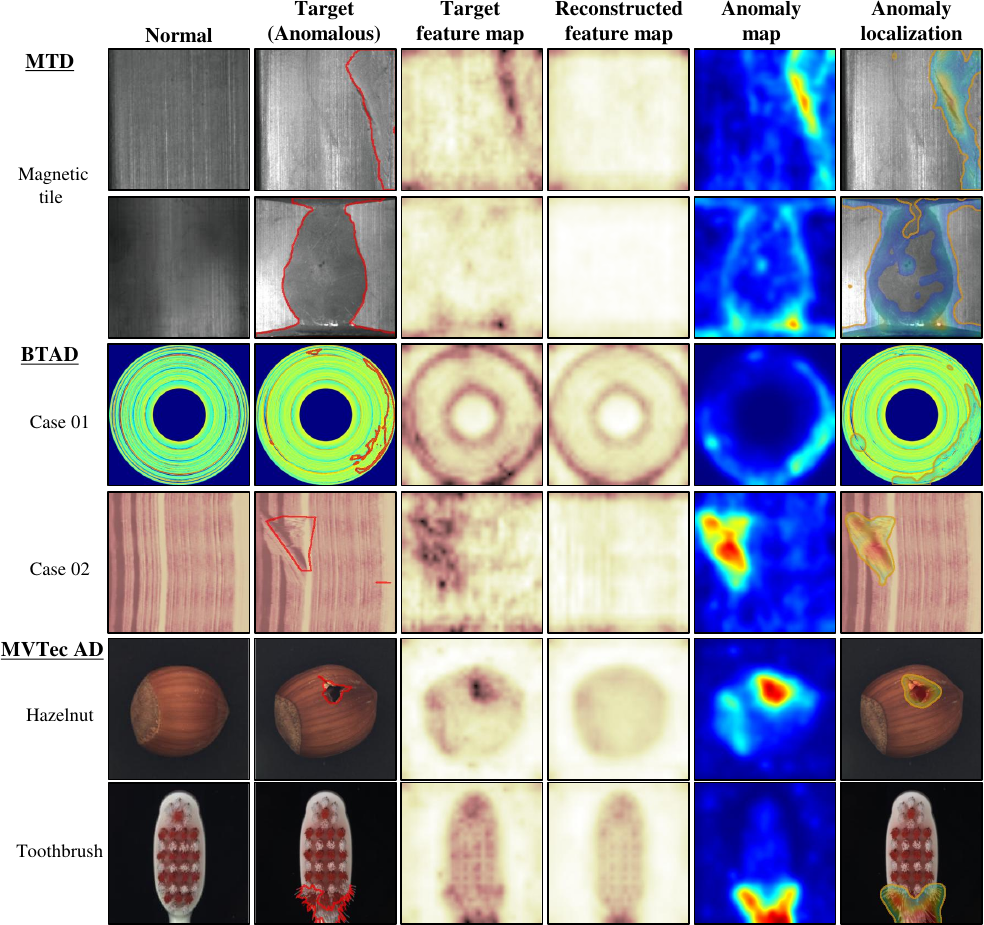}
  \caption{Qualitative results of our method on three benchmark datasets. For each category, we show a nominal image (normal), the target image (abnormal), the target feature map, the reconstructed feature map,  the anomaly map, and the anomaly localization result identified by our method. The boundary in \textcolor{red}{red} denotes the ground-truth anomalous regions.}
  \label{fig:5}
\end{figure*}

\section{Experimental Results and Analysis}
\label{sec:sec4}
In this section, we initially introduce the datasets, provide implementation details, and specify the evaluation metrics employed in our experiments. 
Subsequently, we quantitatively compare our method against the state-of-the-art models on multiple benchmark datasets.
Lastly, we carry out ablation studies from various perspectives to provide a comprehensive analysis of our approach.

\subsection{Implementation Detail}
\label{sec:sec4-1}
Following~\cite{bergmann2020uninformed}, we employ WideResNet50~\cite{zagoruyko2016wide} as the backbone network, which is a widely adopted configuration for method comparison on anomaly localization.
Input images are resized to $256\times256$, and no data augmentation is applied.
STFPM~\cite{wang2021student} uses different networks in the reported paper, and we tried to reproduce the results on WideResNet50 in order to provide a fair and consistent comparison.
Similarly, we retrain DeSTSeg by excluding the data augmentation procedure to maintain consistency across our comparisons.
All experiments are conducted on an RTX~4090 GPU.

For quantitative evaluation metrics, we evaluate our algorithm using the area under the receiver-operator curve (AUROC: \%), the per-region overlap (PRO: \%), and the average precision (AP: \%).
PRO metric uses up to a false positive rate of 30\%, as recommended by~\cite{bergmann2020uninformed}.
These metrics compute the average accuracy of pixel-level localization per category.
The PRO score, which considers the overlap and recovery of connected anomaly components, provides a more comprehensive performance assessment by accounting for variations in anomaly sizes.
The AP score, computed as the area under the precision-recall curve, is more appropriate for samples with small anomalies~\cite{zavrtanik2021draem}.

\subsection{Anomaly Localization on Benchmark Datasets}
\label{sec:sec4-2}
\subsubsection{Magnetic Tile Defect (MTD)~\cite{huang2020surface}}
MTD contains 925 defect-free magnetic tile images and 392 anomalous images with varied illumination levels and image sizes. 
The test set is provided with ground-truth masks.
Following~\cite{rudolph2021same,roth2022towards}, we evaluate 20\% of the defect-free images for testing, while the remaining images are used for training.
We employ feature maps obtained from subsequent hierarchy levels 2, 3, 4, and the last level (where the last level is utilized only for subspace-based sampling). The parameter values are set as follows: $s_{\mathrm{ref}}=10$ and $s=7$.

In Table~\ref{tab:MTD}, we show the results on MTD.
The best result in each row is shown in bold, and the second-best result is underlined.
We can observe that our method outperforms competing approaches across both evaluation metrics, particularly surpassing prior memory bank-based methods.
Typically, memory bank-based methods struggle to deal with anomalies in categories with irregular texture patterns. This well demonstrates the low robustness in performing feature matching for out-of-memory bank features.
Conversely, our approach is less dependent on the memory bank due to its ability to adaptively generate feature representations using subspace-guided feature reconstruction.
The visual results in Fig.~\ref{fig:5} also show that the proposed method achieves robust anomaly localization against light source imbalances.
However, the sample with global anomalies, as shown in the bottom result on MTD, tends to cause False Negative.
Importantly, this issue is not caused by our feature reconstruction approach but rather by the low sensitivity to the overall shape of the CNN as a backbone network.

\subsubsection{BeanTech Anomaly Detection (BTAD)~\cite{mishra2021vt}}
BTAD contains three categories of industrial products, comprising 2,540 samples. The training set comprises only anomaly-free samples, while the test set comprises both anomaly-free and anomalous samples. 
As with the MTD, the test set is provided with ground-truth masks.
We employ feature maps obtained from subsequent hierarchy levels 2, 3, and 4 (where hierarchical level 4 is utilized only for subspace-based sampling). The parameter values are set as follows: $s_{\mathrm{ref}}=80$ and $s=40$.

In Table~\ref{tab:BTAD}, we report the results on BTAD.
The best result in each column is shown in bold, and the second-best result is underlined.
The significant difference between AUROC and PRO scores in this dataset highlights the importance of the PRO metric, which considers the variation in anomaly size, for evaluation compared to other datasets. 
Notably, our method achieves state-of-the-art anomaly localization in the PRO metric. 
Despite its easy implementation using memory bank-based feature reconstruction, our approach outperforms sophisticated methods such as normalizing flow and knowledge distillation-based approaches.
In the visual results in Fig.~\ref{fig:5}, it can be observed that our method yields accurate anomaly localization, even for targets with irregular textures that do not perfectly match the nominal samples.

\subsubsection{MVTec AD~\cite{bergmann2019mvtec}}
MVTec AD, which is widely recognized as a standard benchmark for industrial anomaly localization tasks, consists of 15 categories of industrial products (such as carpet, tile, bottle, etc.), comprising a total of 5,354 images. 
Each category is divided into two sets: a training set that exclusively contains anomaly-free samples and a test set that includes both anomaly-free and anomalous samples. 
The test set contains various defect types associated with specific products, and corresponding ground-truth masks are provided to indicate the location of anomalies within the images.
We employ feature maps obtained from subsequent hierarchy levels 2, 3, and 4  (where hierarchical level 4 is utilized only for subspace-based sampling). The parameter values are set as follows: $s_{\mathrm{ref}}=40$, $s=17$, and $\epsilon=10^{-6}$.

The results for anomaly localization on MVTec AD are listed in Table~\ref{tab:MVT_AUROC} and Table~\ref{tab:MVT_PRO}.
We can confirm that our method is competitive with state-of-the-art methods regarding two evaluation metrics and particularly outperforms reconstruction-based methods AE$_{SSIM}$~\cite{bergmann2019mvtec}, DFR~\cite{yang2020dfr}, and DSR~\cite{zavrtanik2022dsr}.
We assume the reason can be attributed to that the proposed feature reconstruction mechanism is designed to only reconstruct normal features that are faithful to the nominal features, thereby reducing false negatives in anomaly localization.
Specifically, compared to conventional reconstruction methods, modeling the low-dimensional subspace from nominal samples enables more reliable feature approximation to unseen targets that can be represented in the same subspace.
To qualitatively evaluate the reconstructed features of our method, the visual results are shown in Fig.~\ref{fig:5}.
Our method is able to perfectly reconstruct normal patterns even for the subtle appearance features of "Toothbrush" and assign high anomaly scores to the defective regions.

\subsubsection{Overall anomaly localization performance}
Table~\ref{tab:AP} reports the overall anomaly localization performance of each method in terms of AP.
The proposed method achieves strong anomaly localization performance on the MTD and BTAD datasets and outperforms PatchCore on the MVTec AD.
Moreover, the proposed method surpasses state-of-the-art methods regarding the average performance over the three datasets, including memory bank-based methods (SPADE~\cite{cohen2020sub} and PatchCore~\cite{roth2022towards}) and anomaly simulation-based methods (DeSTSeg~\cite{zhang2023destseg} and SimpleNet~\cite{liu2023simplenet}).
Our subspace-guided feature reconstruction achieves high accuracy in industrial product anomaly localization, even without any dataset-specific training.

\begin{figure*}[t]
    \centering
    \subfigure[]{\includegraphics[width=0.4\linewidth]{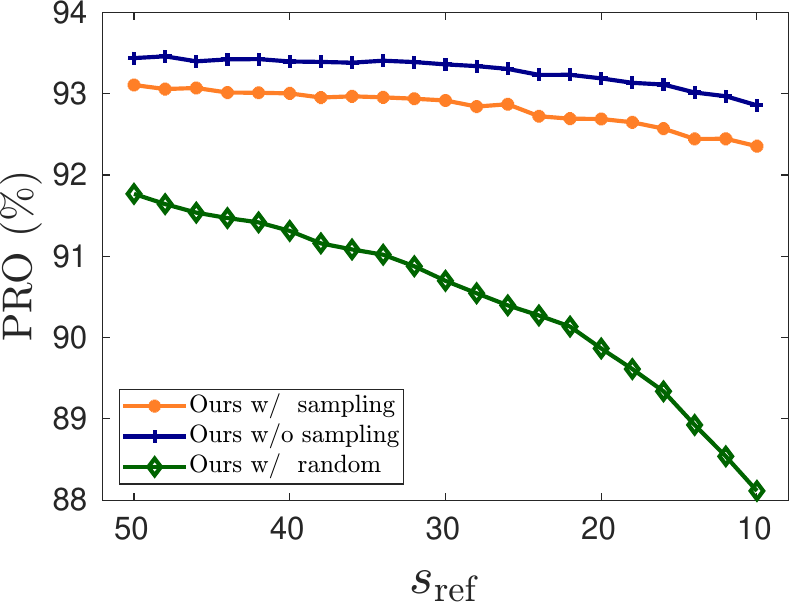}
    \label{fig:6a}
    }
    \subfigure[]{\includegraphics[width=0.4\linewidth]{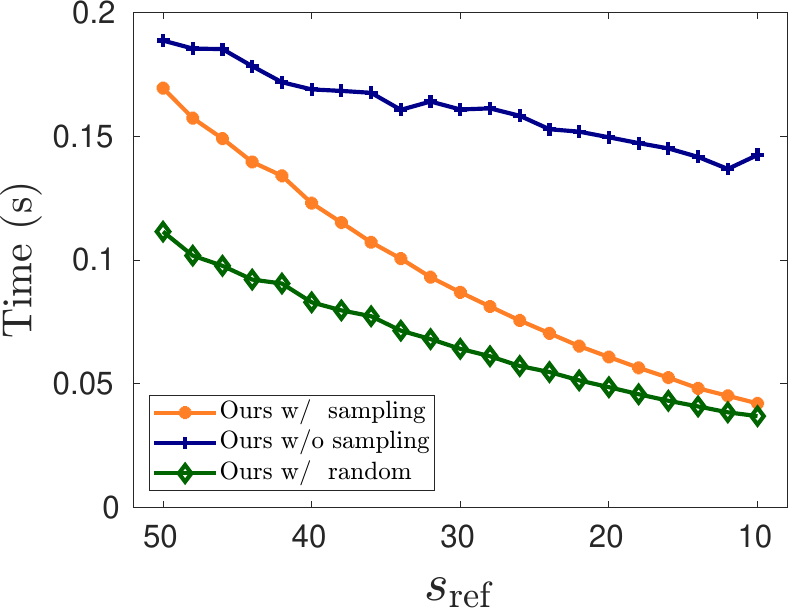}
    \label{fig:6b}
    }
    \caption{Comparison of anomaly localization performance of different sampling mechanisms: (a) PRO and (b) inference time per image.
    }
    \label{fig:6}
\end{figure*}

\subsection{More Evaluations}
\label{sec:sec4-3}
To emphasize the advantages of our proposed model, we conduct a comprehensive analysis to evaluate the effects of our model in the following aspects: (1) Ablation study for the subspace-based sampling technique, (2) Evaluation of the expressiveness of subspace-guided feature reconstruction, (3) Evaluation of the method generalization across different backbones.

\subsubsection{Effectiveness of subspace-based sampling technique}
\label{sec:sec4-3-1}
To further support the computational complexity analysis in Sec.~\ref{sec:sec3_3}, we investigate the anomaly localization performance on MVTec AD by ablating or adopting the proposed sampling technique. 
In addition, we compare the case of replacing the proposed sampling method with simple random sampling.
To confirm the statistical results, we conducted the experiments by varying the number of data stored in the memory bank ($s_\mathrm{ref}$) from 50 to 10, and the parameter determining the sparsity in the reconstruction step is fixed at $s = s_\mathrm{ref}/2$. 
The reported inference times encompass the forward pass through the backbone network.
As shown in Fig.~\ref{fig:6}\subref{fig:6a}, we can confirm that the PRO score of the proposed sampling technique consistently stays within a $0.5$\% range of that achieved without the sampling technique, regardless of the value of $s_\mathrm{ref}$.
In particular, Fig.~\ref{fig:6}\subref{fig:6b} shows that when $s_\mathrm{ref}$ is less than 20, the PRO score drops significantly for the random sampling method, while the proposed sampling method maintains the PRO score, and the inference time remains comparable to that of the random sampling.

\begin{figure}[t]
\centering
    \includegraphics[width=0.8\linewidth]{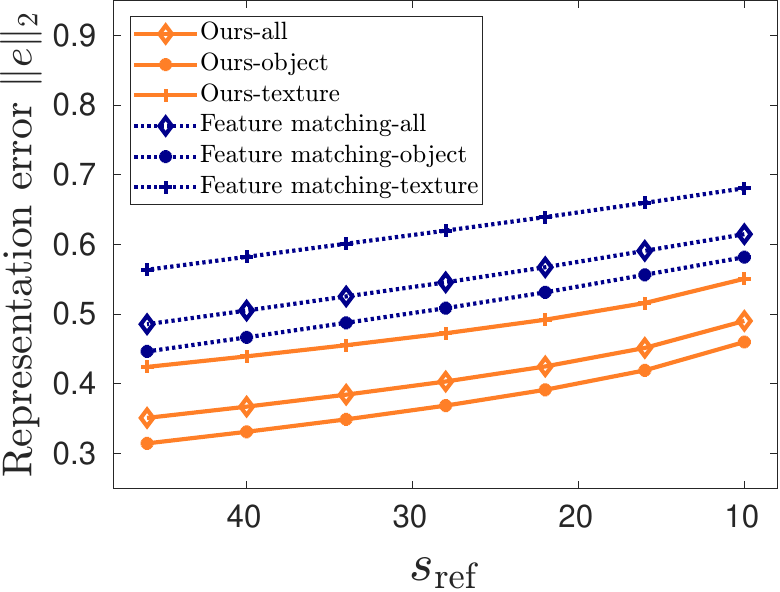}
  \caption{Evaluation of data coverage for out-of-bank data: representation error of out-of-bank data as a function of $s_{\mathrm{ref}}$ in the three settings of object, texture, and both in MVTec AD dataset~\cite{bergmann2019mvtec}.}
  \label{fig:7}
\end{figure}

\begin{figure}[t]
\centering
    \includegraphics[width=0.8\linewidth]{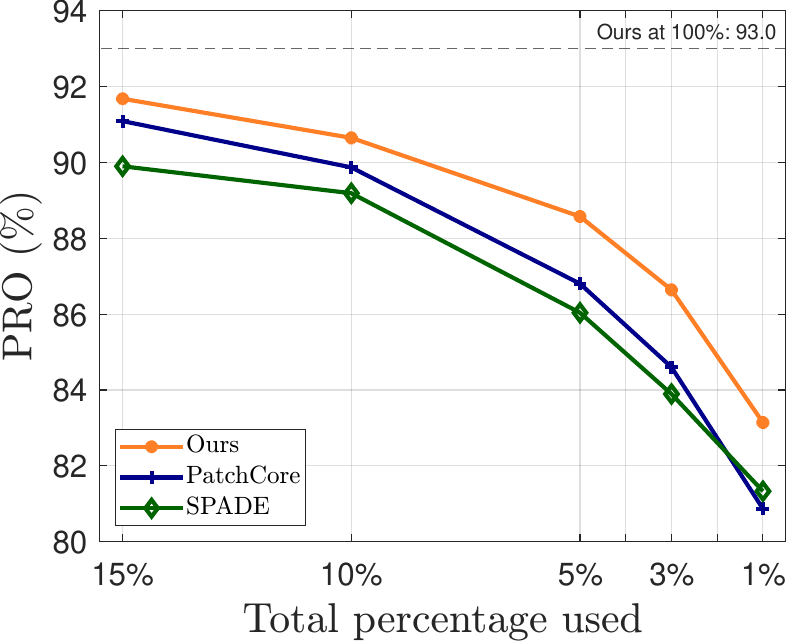}
  \caption{Curves of the change of PRO score with respect to the total percentage used in the overall nominal training data under low-shot configuration on MVTec AD dataset~\cite{bergmann2019mvtec}.}
  \label{fig:8}
\end{figure}

\subsubsection{Expressiveness of subspace-guided feature reconstruction.}
\label{sec:sec4-3-2}
In approaches based on the memory bank framework, while storing all nominal data in a memory bank can enhance anomaly localization performance, the computational complexity poses a significant impediment, rendering it impractical.
Thus, the performance of such an approach depends on the ability to alternatively mimic out-of-bank features with limited in-sample features, even to encourage a sufficient data coverage.
To verify the data coverage for out-of-bank data, we conduct a comparative analysis of the representation error $\|\boldsymbol{e}\|_{2}$ of out-of-bank features with the proposed reconstruction methodology and the feature matching employed by SPADE~\cite{cohen2020sub} and PatchCore~\cite{roth2022towards} as follows:
\begin{equation}
    \|\boldsymbol{e}\|_{2} = \frac{1}{N}\sum_{j \in \{1,\cdots,N\}} \|\boldsymbol{x}_{j} - \boldsymbol{x}_{j}^{\ast}\|_{2},
    \label{eq:eq9}
\end{equation}
where $\boldsymbol{x}_{j}^{\ast}$ approximates or is the nearest vector to $\boldsymbol{x}_{j}$.
In our reconstruction methodology, $\boldsymbol{x}_{j}^{\ast}$ denotes the data point that is represented through the linear combination in Eq.~(\ref{eq:eq8}). 
In the feature matching-based method, the top-$s_{\mathrm{ref}}$ data that exhibit proximity in spatial distance to $\boldsymbol{x}_{j}$ are stored in $M$, and $\boldsymbol{x}_{j}^{\ast}$ denotes the nearest neighbor data point as follows: $\boldsymbol{x}_{j}^{\ast}=\min_{\boldsymbol{x}_{i} \in \mathcal{M}} \|\boldsymbol{x}_{j} - \boldsymbol{x}_{i}\|_{2}$.
In Fig.~\ref{fig:7}, we can observe that our subspace-guided feature reconstruction in the three settings in MVTec AD provides a better representation of out-of-bank data than the feature matching-based method.
Particularly in the texture category, the feature patterns of the out-of-bank data cannot be well represented by feature matching because they are diverse for the data coverage of the in-sample data.
Conversely, our approach is able to mimic the out-of-bank features using limited data by the benefit of the self-expressive model.
This result supports the concept that our methodology enables adaptive feature representation even with a limited number of in-sample data, which leads to improved anomaly localization performance for the texture categories presented in Sec.~\ref{sec:sec4-2}.

We also investigate the expressiveness of our feature reconstruction methodology on limited nominal data. 
In particular, we vary the amount of training samples from 15\% to 1\% of the total nominal training data on the MVTec AD dataset and compare it to memory bank-based approaches using WideResNet50. 
As shown in Fig.~\ref{fig:8}, even when utilizing only 15\% of the total nominal training data, our method remains within a 1.5\% range of the performance compared to the case of fully using the training data (Fig.~\ref{fig:8}, dashed line). 
Furthermore, we can see that our method consistently outperforms all competitive methods when the number of training samples is reduced to 1\%. 
We can thus confirm that our method suffices to present rich expressiveness even on a limited number of nominal training data.

\begin{figure}[t]
\centering
    \includegraphics[width=0.9\linewidth]{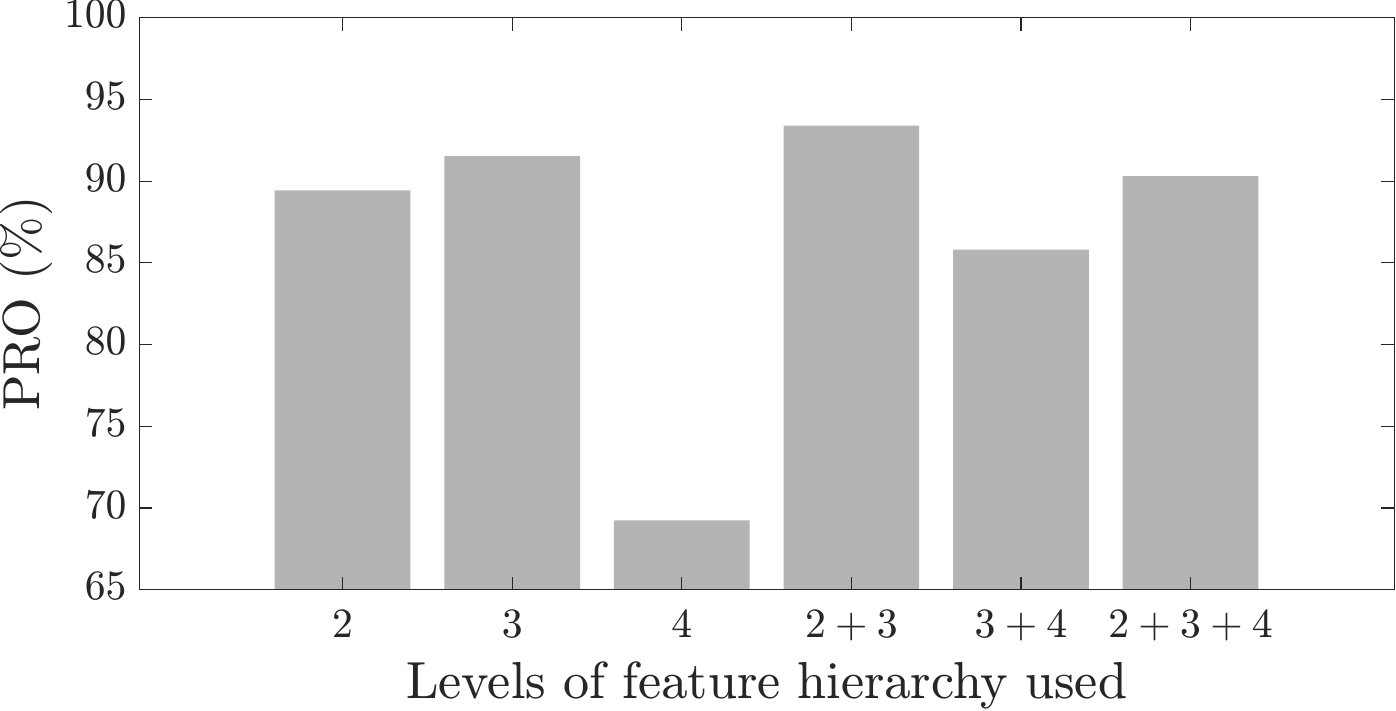}
  \caption{Evaluation of network feature hierarchies and anomaly localization performance in terms of PRO on MVTec AD dataset~\cite{bergmann2019mvtec}.}
  \label{fig:9}
\end{figure}

\subsubsection{Method generalization across different hierarchies and backbones}
\label{sec:sec4-3-3}
We investigate the generalization capability of the proposed reconstruction methodology (Sec.~\ref{sec:sec3_1}) for features stored in the memory bank.
Fig.~\ref{fig:9} shows the single or multi-scale localized predictions obtained by indexing WideResNet50 blocks, where $l\in\{2,3,4\}$.
Obviously, the result using the features from hierarchy level 4 is significantly lower than those using other features.
This is partly due to the fact that high-level features, which are very deep and abstract, lose localized information due to their low resolution.
Moreover, these features in ImageNet pre-trained networks are biased toward natural image classification tasks~\cite{roth2022towards}.
Consequently, a practical choice would involve utilizing at least a combination of low-level features with a high resolution in our feature reconstruction methodology.

Table~\ref{tab:ablation1} shows the performance of the proposed method (with and without a sampling technique) using different backbones. 
We observe that the anomaly localization performance with the proposed approach remains consistently stable across different backbones. 
In terms of the timings, we can see that our sampling method significantly improves the inference time on the ResNet family, while the gain on the EfficientNet family is comparably small. 
This is expected to be due to the small difference in the dimensionality of feature maps at successive hierarchical levels, which causes the cost for sampling to dominate the overall computational expense.

\begin{table}[t]
\centering
\caption{Anomaly localization performance for different backbones on MVTec AD dataset~\cite{bergmann2019mvtec}. Comparative results are measured by AUROC/PRO/inference time per image (s).}
\label{tab:ablation1}
\vspace{5pt}
\scalebox{0.95}{
\begin{tabular}{lcccc}
\toprule
Backbone      & Ours (w/o sampling)  & Ours (w/  sampling)\\ \hline
\textbf{ResNet50}~\cite{he2016deep}             & 97.3 / 92.3 / 0.180 & 97.2 / 91.8 / 0.115 \\
\textbf{ResNet101}~\cite{he2016deep}            & 97.6 / 92.9 / 0.183 & 97.5 / 92.5 / 0.116 \\
\textbf{WideResNet50}~\cite{zagoruyko2016wide}  & 97.6 / 93.4 / 0.184 & 97.6 / 93.0 / 0.117\\
\textbf{WideResNet101}~\cite{zagoruyko2016wide} & 97.8 / 93.7 / 0.183 & 97.7 / 93.3 / 0.123 \\
\textbf{EfficientNet-B1}~\cite{tan2019efficientnet} & 97.5 / 93.2 / 0.068 & 97.3 / 92.9 / 0.065 \\
\textbf{EfficientNet-B7}~\cite{tan2019efficientnet} & 97.3 / 93.2 / 0.106 & 97.1 / 92.9 / 0.084 \\
\bottomrule
\end{tabular}
}
\end{table}

%% file: paper/conclusion.tex
\section{Conclusion}
\label{sec:sec5}
In this paper, we proposed a novel subspace-guided feature reconstruction framework for anomaly localization.
Moreover, we proposed a subspace-based sampling technique that aggregates the basis vectors of forming a subspace to retain high inference speeds.
Extensive experiments on several public datasets have validated the efficiency and effectiveness of our proposed method for localization accuracy and dependence on the number of data.

Although our method is designed to efficiently localize anomalies using subspace-based sampling, its reliance on the memory bank may induce some computational overhead imposed by the calling procedure.
In the future, we would like to devise a memory bank-free yet end-to-end trainable reconstruction model using unsupervised self-expressive networks. In addition, our method currently assumes the data is distributed in linear subspaces. We would also like to focus on developing a stronger reconstruction mechanism to explore the non-linear subspaces for anomaly localization.